\documentclass[runningheads]{llncs}
\usepackage[T1]{fontenc}
\usepackage{graphicx}
\usepackage{color}

\usepackage{enumitem} 
\usepackage{todonotes}
\usepackage{xspace}
\usepackage{booktabs}
\usepackage{multirow}
\usepackage{url}
\usepackage{amsmath}
\usepackage{amssymb}
\usepackage{subfig}
\usepackage{tablefootnote}
\usepackage{colortbl}
\usepackage{newtxtext}
\usepackage[normalem]{ulem}
\usepackage{refcount}
\usepackage{hyperref}

\newcommand{\act}[1]{\texttt{#1}}
\newcommand{\attr}[1]{\textsl{#1}}

\newcommand{\bpi}{\textsc{BPI2012}}
\newcommand{\finel}{\textsc{Fines2015}}
\newcommand{\loans}{\emph{Loans}}
\newcommand{\fines}{\emph{Fines}}
\newcommand{\loanreq}{\emph{Loan request handling}\xspace}
\newcommand{\traffine}{\emph{Traffic fine management}\xspace}

\newcommand{\simeval}{simulation evaluation\xspace}
\newcommand{\logeval}{test log evaluation\xspace}

\newcommand{\random}{\textsc{Random}\xspace}
\newcommand{\optimal}{\textsc{Optimal}\xspace}
\newcommand{\usual}{\textsc{Customary}\xspace}
\newcommand{\splitA}{60\%-40\%}
\newcommand{\splitB}{80\%-20\%}
\newcommand{\allt}{\textsc{All}\xspace}
\newcommand{\optimalt}{\textsc{Optimal P.}\xspace}
\newcommand{\notoptimalt}{\textsc{Non-Optimal P.}\xspace}

\newcommand{\action}{\emph{action}}
\newcommand{\state}{\emph{state}}
\newcommand{\rew}{\emph{reward}}

\newcommand{\la}{\textsc{LA}}
\newcommand{\hf}{\textsc{HF}}
\newcommand{\ef}{\textsc{EF}}

\definecolor{Gray}{gray}{0.9}

\begin{document}
\title{Learning to act: \\ a Reinforcement Learning approach to recommend the best next activities}
%
%
\titlerunning{Learning to act}
%
%
\author{Stefano Branchi\inst{1} \and Chiara Di Francescomarino\inst{1} \and Chiara Ghidini\inst{1} \and \\  David Massimo\inst{2}  \and Francesco Ricci\inst{2}  \and Massimiliano Ronzani\inst{1}}
\authorrunning{Branchi et al.}
%
\institute{Fondazione Bruno Kessler, Trento, Italy
\email{sbranchi|dfmchiara|ghidini|mronzani@fbk.eu} \and
Free University of Bozen-Bolzano, Bolzano, Italy
\email{David.Massimo|fricci@unibz.it}}
%
\maketitle              
\begin{abstract}

The rise of process data availability has recently led to the development of data-driven learning approaches. However, most of these approaches restrict the use of the learned model to predict the future of ongoing process executions.
The goal of this paper is moving a step forward and leveraging available data to \emph{learning to act}, by supporting users with recommendations derived from an optimal strategy (measure of performance). We take the optimization perspective of one process actor and we recommend 
the best activities to execute next,
in response to what happens in a complex external environment, where there is no control on exogenous factors.
To this aim, we investigate an approach that learns, by means of Reinforcement Learning, the optimal policy from the observation of past executions and recommends the best activities to carry on for optimizing a Key Performance Indicator of interest.
The validity of the approach is demonstrated on two scenarios taken from real-life data. 
%

\end{abstract}
\section{Introduction}
\label{sec:intro}

In the last few years, a number of works have proposed approaches, solutions and benchmarks in the field of Predictive Process Monitoring~\cite{DBLP:conf/caise/MaggiFDG14,DBLP:conf/bpm/Francescomarino18}. Predictive Process Monitoring leverages the analysis of historical execution traces in order to predict the unrolling of a process instance that has been only partially executed.
%
However, most of these efforts 
have not used the predictions to explicitly support user with recommendations, i.e., with a concrete usage of these predictions. In fact, there is a clear need of \emph{actionable} process management systems~\cite{DBLP:journals/corr/abs-2201-12855} able to support the users with recommendations about the best actions to take.

The overall goal of this paper is therefore moving a step forward, towards the implementation of a \emph{learning to act} system, in line with the ideas of Prescriptive Process Monitoring~\cite{fahrenkrog2019fire,metzger2020triggering}.
Given an ongoing business process execution, Prescriptive Process Monitoring aims at recommending activities or interventions with the goal of optimizing a target measure of interest or Key Performance Indicator (KPI). 
State-of-the-art works have introduced methods for raising alarms or triggering interventions, to prevent or mitigate undesired outcomes, as well as for recommending the best resource allocation. Only few of them have targeted the generation of recommendations of the next activity(ies) to optimize a certain KPI of interest~\cite{weinzierl2020predictive,groger2014prescriptive,de2020design}, such as, the cycle time of the process execution. 
Moreover, none of them explicitly considers the process execution in the context of a complex environment that depends upon exogenous factors, including how the other process actors behave. In this setting, identifying the best strategy to follow for a target actor, is not straightforward. 

In this paper, we take the perspective of one target actor and we propose a solution based on Reinforcement Learning (RL): to recommend to the actor what to do next in order to optimize a given 
KPI of interest for this actor.
To this aim, we first learn, from past executions, the response of the environment (actions taken by other actors) to the target actor's actions, 
and we then leverage RL to recommend the best activities/actions to carry on to optimize the KPI. 


In the remainder of the paper after introducing some background concepts (Section~\ref{sec:background}), we present two concrete Prescriptive Process Monitoring problems that we have targeted (Section~\ref{sec:running}). Section~\ref{sec:problem} shows how a Prescriptive Process Monitoring problem can be mapped into RL, while  
Section~\ref{sec:evaluation} applies the proposed RL approach to the considered problems and evaluates its effectiveness. Finally, Section~\ref{sec:related} and Section~\ref{sec:conclusion} present related works and conclusions, respectively.

\section{Background}
\label{sec:background}


\subsection{Event logs}
\label{ssec:log}
An \emph{event log} consists of traces representing executions of a process (a.k.a.~a case).
A trace is a sequence of \emph{events}, each referring to the execution of an activity (a.k.a.~an event class). Besides timestamps, indicating the time in which the event has occurred, events in a trace may have a data payload consisting of attributes,  such as, the resource(s) involved in the execution of an activity, or other data recorded during the event. Some of these attributes do not change throughout the different events in the trace, i.e., they refer to the whole case (\emph{trace attributes}); for instance, the personal data (\attr{Birth date}) of a customer in a loan request process. 
Other attributes are specific of an event (\emph{event attributes}), for instance, the employee who creates an offer (\attr{resource}), which is specific of the activity \act{Create offer}.

\subsection{Prescriptive Process Monitoring}
\label{ssec:prescriptive}
Prescriptive Process Monitoring~\cite{fahrenkrog2019fire,metzger2020triggering} is a branch of Process Mining that aims at suggesting activities or triggering interventions for a process execution for optimizing a desired Key Performance Indicator (KPI). 
Differently from Predictive Process Monitoring approaches, which aim at predicting the future of an ongoing execution trace, Prescriptive Process Monitoring techniques aim at recommending the best interventions for achieving a target business goal.
%
For instance, a bank could be interested in minimizing the cost of granting a loan to a customer. In such a scenario, the KPI of interest for the bank is the cost of the activities carried out by the bank's personnel in order to reach an agreement with the customer. The best actions that the bank should carry out to achieve the business goal (reaching the agreement while minimizing the processing time) can be recommended to the bank.


\subsection{Reinforcement Learning}
\label{ssec:reinforcement}

Reinforcement Learning (RL)~\cite{Sutton1998,junyanhu} refers to techniques providing an intelligent agent the capability to act in an environment, while maximizing the total amount of reward received by its actions. 
At each time step $t$, the agent chooses and executes an \emph{action} $a$ in response to the observation of the \emph{state} of the environment $s$. 
The action execution causes, at the next time step $t+1$, the environment to stochastically move to a new state $s'$, and gives the agent a \emph{reward} $r_{t+1}=\mathcal{R}(s,a,s')$ that indicates how well the agent has performed.
The probability that, given the current state $s$ and the action $a$, the environment moves into the new state $s'$ 
is given by the state transition function $\mathcal{P}(s,a,s')$. 
The learning problem is therefore described as a discrete-time Markov Decision Process (MDP), 
%
%
%
which is formally defined by a tuple $M=(\mathcal{S}, \mathcal{A}, \mathcal{P},\mathcal{R}, \gamma)$:
\begin{itemize}[label=$\bullet$]
    \item $\mathcal{S}$ is the set of states.
    
    \item $\mathcal{A}$ is the set of agent's actions.
    
    \item $\mathcal{P}:\mathcal{S} \times \mathcal{A} \times \mathcal{S} \rightarrow [0,1]$ is the transition probability function. $\mathcal{P}(s,a,s') = Pr(s_{t+1} = s' | s_t = s, a_t = a)$ is the probability of transition (at time $t$) from state $s$ to state $s'$ under action $a \in \mathcal{A}$.
   
    \item $\mathcal{R}: \mathcal{S} \times \mathcal{A} \times \mathcal{S} \rightarrow \mathbb{R}$ is the reward function. $\mathcal{R}(s,a,s')$ is the immediate reward obtained by the transition from state $s$ to $s'$ with action $a$.
    
    \item $\gamma \in [0,1]$ is a parameter that measures how much the future rewards are discounted with respect to the immediate reward.
    Values of $\gamma$ lower than 1 model a decision maker that discount the reward obtained in the more distant future.\footnote{In this paper we set $\gamma=1$, hence equally weighting the reward obtained at each action points of the target actor.} 
\end{itemize}

An MDP satisfies the \emph{Markov Property}, that is, given $s_t$ and $a_t$, the next state $s_{t+1}$ is conditionally independent from all prior states and actions and it only depends on the current state, i.e., $Pr(s_{t+1}| s_t, a_t)=Pr(s_{t+1}|s_0, \cdots, s_t, a_0, \cdots, a_t)$.

%
The goal of RL is computing a \emph{policy} that allows the agent to maximize the cumulative reward. A policy $\pi: \mathcal{S} \rightarrow \mathcal{A}$ is a mapping from each state $s \in \mathcal{S}$  to an action $a \in \mathcal{A}$,
and the \emph{cumulative reward} is the (discounted) sum of the rewards obtained by the agent while acting at the various time points $t$.
%
%
The value of taking the action $a$ in state $s$ and then continuing to use the policy $\pi$, is the expected discounted cumulative reward of the agent, and it is given by the \emph{state-action value function}:   
$Q^{\pi}(s,a) = \mathbb{E}_\pi
 (\Sigma_{k=0}^\infty\gamma^k r_{k+t+1} |s=s_t,a=a_t )$, where $r_{t+1}$ is the reward obtained at time $t$.
The \textit{optimal} policy $\pi^*$ dictates to a user in state $s$ to perform the action that maximises $Q(s,\cdot)$. Hence, the optimal policy $\pi^*$ maximises the cumulative reward that the user obtains by following the actions recommended by the policy $\pi^*$.
%
%
%
Action-value functions can be estimated from experience, e.g., by averaging the actual returns for each state (action taken in that state), as with \emph{Monte Carlo methods}. 


Different algorithms can be used in RL~\cite{Sutton1998}. 
Among them we can find the \emph{value} and the \emph{policy iteration} approaches. In the former the optimal action-value function $Q^{*}(s,a)$  is obtained by iteratively updating the estimate $Q^{\pi}(s,  a)$. In the latter, the starting point is an arbitrary policy $\pi$ that is iteratively evaluated (\emph{evaluation phase}) and improved (\emph{optimization phase}) until convergence. Monte Carlo methods are used in the policy evaluation phase for computing, given a policy $\pi$, for each state-action pair ${(s,a)}$, the action-value function $Q^{\pi}(s,a)$. The estimate of the value of a given state-action pair $(s,a)$ can be computed by averaging the sampled returns that originated from $(s,a)$ over time. Given sufficient time, this procedure can construct a precise estimate $Q$ of the action-value function $Q^{\pi}$. 
%
%
In the policy improvement step, the next policy is obtained by computing a greedy policy with respect to $Q$: given a state $s$, this new policy returns an action that maximizes $Q(s,\cdot)$. 

\section{Two Motivating Scenarios}
\label{sec:running}

We introduce here the considered problem by showcasing two real processes that involve one target actor, whose reward is to be maximised, and some more actors, contributing to determine the outcome of the process (environment).

\paragraph{Loan request handling} (\loans). 
In a financial institute handling loan requests, customers send loan request applications and the bank decide either to decline an application, or to request further details to the customer, or to make an offer and start a negotiation with the customer. During the negotiation phase, the bank can contact the customer and possibly change its offer to encourage the customer to finally accept the bank's offer.

The bank  aims at maximizing its payoff by trying to sign agreements with the customer, while reducing the costs of the negotiation phase, i.e., stopping negotiations that will not end up with an agreement. The bank is therefore interested to implement the best strategy to follow (actions) in order to maximize its interest.

\paragraph{Traffic fine management} (\fines). 
In a police department in charge to collect road traffic fines, as in the scenario presented in~\cite{DBLP:journals/computing/MannhardtLRA16}, fines can be paid (partly or fully) either immediately after the fine has been issued, or after the fine notification is sent by the police to the offender’s place of residence, or when the notification is received by the offender. If the entire amount is paid, 
the fine management process is closed.
After being notified by post, the offender can appeal against the fine through a judge and/or the prefecture. If the appeal is successful, the case ends. 

In such a setting, the police department aims at collecting the payment of the invoice by the offender as soon as possible, so as to avoid money wastes due to delays in payments or the involvement of the prefecture/judge. The department indeed receives credits for fast payments, no credits for payments never received and discredits for incorrect fines. The department is therefore interested to receive best action recommendations to maximize the received credits.
\section{Mapping PPM to RL}
\label{sec:problem}

We would like to support a target actor of interest in a process, 
such as, the financial institute or the police department (see Section~\ref{sec:running}), by providing them with recommendations for the best activities to execute in order to maximize their profit and their credits, respectively. To this aim, we leverage RL, by transforming the PPM problem of recommending the next activities to optimize a given KPI, into an RL problem, where the \emph{agent} is the actor we are supporting in the decision making (e.g., the bank or the police department), and the \emph{environment} is represented by the external factors---especially
 the activities carried out by the other actors involved in the process execution (e.g., the customer or the offender). 
We define our MDP so that:
\begin{itemize}[label=$\bullet$,topsep=0.3pt]
    
    \item an \emph{action}, to be recommended, is an activity of the actor of interest (\emph{agent}) (e.g., the bank activity \act{Create offer});
    
    \item a \emph{state} is defined by taking into account 
    the following variables:
		\begin{itemize}[label=$-$]
		\item the last activity executed by the actor of interest (e.g., the creation of a new offer by the bank) or by the other actors defining the stochastic response of the environment (e.g., the bank offer acceptance by the customer);
	
		\item some relevant information about the history of 
		the execution (e.g., the number of phone calls between the bank and the customer);
		
		\item other aspects 
		defining the stochastic response of the environment (e.g., the amount of the requested loan);
        \end{itemize} 
		
    A \state{} is hence represented by a tuple $\langle \la, \hf, \ef \rangle$, where \la{} is the last activity executed by the actor of interest or by one of the other actors involved in the process, \hf{} is a vector of features describing some 
	relevant information of the process execution history 
	and \ef{} is a vector of features further describing the environment response to the actions of the actor of interest.
    
    \item the \emph{reward function} is a numerical value that 
     transforms the KPI of interest, computed on the complete execution, in a 
     utility function at the level of single action.
\end{itemize}

\begin{figure}[t]
  \centering
  \includegraphics[width=\textwidth]{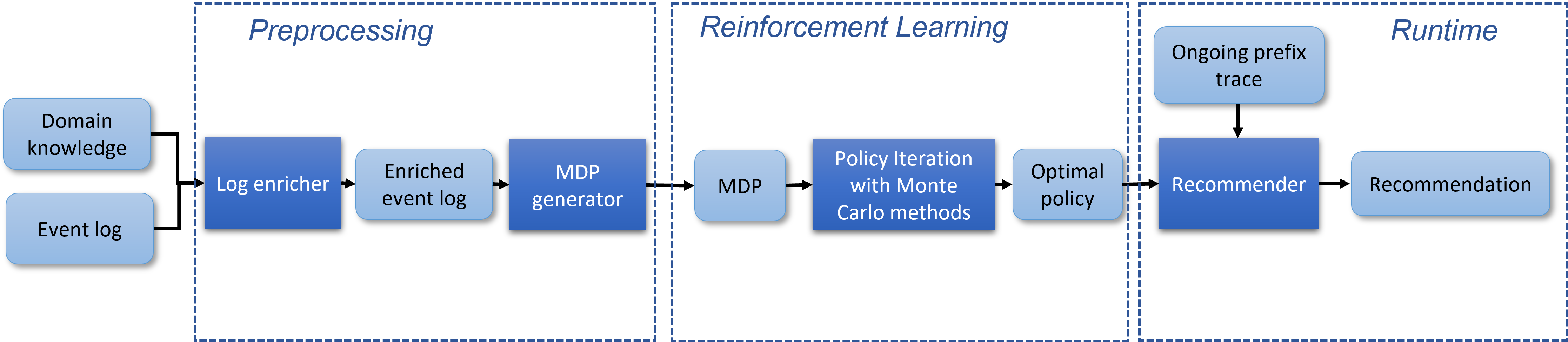}
  \caption{Architecture of the RL solution}
  \vspace{-0.5cm}
  \label{fig:architecture}
\end{figure}

Actions, states and reward function can be defined for each specific problem by leveraging the information contained in the event log and some domain knowledge. The activities we are interested to recommend and those describing the stochastic response  of the environment can be extracted from the event log.
The relevant information about the history of 
the process execution can also be extracted from the event log, with some domain-specific pre-processing (e.g., counting the number of phone calls between the bank and the customer). The stochastic responses of the environment to the actor's actions can also be mined from the event log through trace attributes (e.g., the amount of the requested loan). 
Finally, information contained in event logs can be used to estimate the reward function for each state transition and action (e.g., in case the reward function is related to the process/event cycle time, the average duration of events of a certain type can be used to estimate the reward of a given state).
Figure~\ref{fig:architecture} shows the architecture of the RL-based solution designed to solve the problem of recommending the next best activities to optimize a certain KPI. 
%
The input is an event log containing historical traces related to the execution of a process, and some domain knowledge, specifying the KPI of interest and the information that allows for the identification of actions, states and of the reward function. There are three main processing phases: 

\begin{itemize}[label=$\bullet$,topsep=0.3pt]

\item \emph{preprocessing phase:} the event log is preprocessed in order to learn a representation of the environment (i.e., the MDP). First, the event log is cleansed and the domain knowledge leveraged in order to annotate it.
In detail, the event log is 
(i) filtered in order to remove low-frequency variants (with occurrence frequency lower than 10\%) and activities that are not relevant for the decision making problem;
(ii) enriched with attributes obtained by aggregating and preprocessing information related to the execution; 
(iii) annotated by specifying the agent's activities to be considered as actions; attributes and environment activities to be used for the state definition;  
attributes to be used for the computation of the reward function. 

Once the event log has been enriched and annotated, it can be used for building the MDP that defines the RL problem. 
To this aim, we start from the scenario-specific definition of action and state and, by replaying the traces in the event log, we build a directed graph, where each node corresponds to a state and each edge is labelled with the activity allowing to move from one node state to the other. Moreover, for each edge, the probability of reaching the target node (computed based on the number of traces in the event log that reach the corresponding state) and the value of the reward function are computed. Each edge is hence mapped to the tuple $(s, a, s', \mathcal{P}(s'|s,a), \mathcal{R})$ where $s$ is the state corresponding to the source node of the edge, $a$ is the action used for labelling the edge, $s'$ is the state corresponding to the target node of the edge, $\mathcal{P}(s'|s,a)$ is computed as the percentage of the traces that reach the state $s'$ among the traces that reach state $s$ and execute $a$, and $\mathcal{R}$ is the value of the reward function.



\item \emph{RL phase:} the RL algorithm is actually applied to compute the optimal policy $\pi^\ast$; in this paper we used policy iteration
with Monte Carlo methods.

\item \emph{runtime phase:} given an empty or ongoing execution trace, the policy is queried by the recommender system to return the best activities to be executed next.
\end{itemize}

\section{Evaluation of the Recommendation Policy}
\label{sec:evaluation}

We investigate the capability of the proposed 
approach to recommend
the process activities that allow the target actor to maximize a KPI of interest, i.e., the optimal policy $\pi^*$,
(i) when no activity has been executed yet, that is, the whole process execution is recommended; (ii) at different time steps of the process execution (i.e., at different prefix lengths), that is, when only a (remaining) part of the process execution is recommended. 
We hence explore the following research questions:

\begin{enumerate}[label=\textbf{RQ\arabic*},leftmargin=2\parindent,topsep=1.5pt]

\item\label{RQ1} How does the recommended sequence of  activities (suggested by the optimal policy $\pi^*$) perform in terms of the KPI of interest when no activity has been executed yet?

\item\label{RQ2} How does the recommended sequence of  activities (suggested by the optimal policy $\pi^*$) perform in terms of the KPI of interest at a given point of the execution?
\end{enumerate}

Unfortunately, the complexity of evaluating recommendations in the Prescriptive Process Monitoring domain is well known~\cite{DBLP:conf/bpm/Dumas21}. It relates to the difficulty to estimate the performance of recommendations that have possibly not been followed in practice. In order to answer our research questions, we therefore approximate the value of the KPI of interest (i) by leveraging a simulator (\simeval); (ii) by looking at similar executions in the actual event log  (\logeval). In the next subsections we describe the dataset (Section~\ref{ssec:datasets}), we detail the experimental setting (Section~\ref{ssec:setting}), and we finally report the evaluation results (Section~\ref{ssec:results}).
 

\subsection{Datasets}
\label{ssec:datasets}
We have used two real-world publicly-available datasets that, describing the behaviour of more than one actor, allow us to take the perspective of one of them (target): the BPI Challenge 2012 event log~\cite{vandongen_2012} (\bpi) and the Road Traffic Fine Management event log~\cite{deleoni_mannhardt_2015} (\finel).

The BPI Challenge 2012 dataset relates to a Dutch Financial Institute. The process executions reported in the event log refer to an application process for personal loan (see the 
\loans{} scenario in Section~\ref{sec:running}). 
In this scenario we want to optimize the profit of the bank (\emph{agent}), i.e., to minimize the cost $C$ of granting a loan  to a customer (\emph{environment}) while maximizing the interest $I$ of the bank granting the loan. 
To this aim, we define the KPI of interest for a given execution $e$ as the difference between the amount of interest (if the bank offer is accepted and signed by the customer, namely if the activity \act{Offer accepted} occurs in the trace) and the cost of the employees working time, that is, the value of the KPI for the execution $e$ is $\text{KPI}_{\bpi}(e) = I(e) - C(e)$. The amount of interest depends on the amount class of the loan request: low ($ \text{amount} \le 6000$), medium ($6000 < \text{amount} \le 15000$) and high ($\text{amount} > 15000$). For the low class, the  average interest rate is 16\%, for the medium class, the average interest rate is 18\%, while for the high class is 20\%.
\footnote{The information on the average interest rate is extracted from the BPI2017~\cite{vandongen_2017} dataset which contains data from the same financial institution.}  The cost of the employees' working time is computed assuming an average salary of $18$ euros/h.\footnote{We estimate the average salary of a bank employed in the Netherlands from https://www.salaryexpert.com/salary/job/banking-disbursement-clerk/netherlands.}

\begin{table}[t]
	\centering
	\scalebox{0.8}{%
		\begin{tabular} {l@{\hskip 0.2in}c@{\hskip 0.2in}c@{\hskip 0.2in}c@{\hskip 0.2in}c@{\hskip 0.2in}c}
		\toprule
			\textbf{Dataset} & \textbf{Trace \#} & \textbf{Variant \#} & \textbf{Event \#} &  \textbf{Event class \#} &
			\textbf{Avg.~trace length}\\ \hline
			\bpi & 13087 & 4366 & 262200 & 36 & 20\\
			\finel & 150370 & 231 & 561470 & 11 & 5\\
			\bottomrule
		\end{tabular}
		}
	\vspace{0.3cm}
	\caption{Dataset Description}
	\vspace{-1 cm}
	\label{tab:dataset}
\end{table}

The second dataset collects data related to an information system of the Italian police. The information system deals with the management of road traffic fines procedures, starting from the fine creation, up to the potential offender's appeal to the judge or to the prefecture (see the \fines{}
scenario described in Section~\ref{sec:running}).
Here, we want to maximize the credits received by the police department (\emph{agent}) based on the fine payments received by the offender (\emph{environment}). The department receives $3$, $2$ or $1$ credits if the fines are fully paid within 6, within 12 months, or after 12 months respectively; it does not receive any credits if the fine is not fully paid, while it receives a discredit if the offender appeals to a judge or to the prefecture and wins, 
since these cases correspond to a money waste of the police authority. The KPI value for the execution $e$ is KPI$_{\finel}(e)$, corresponding to the number of credits received for the execution.

Table~\ref{tab:dataset} shows the number of traces, variants, events, event classes and average trace length of the considered datasets.
%
Table~\ref{tab:scenarios} illustrates the MDP components for the two scenarios: the main MDP \emph{actions}; the main MDP \state{} components, i.e., the last activity (\la), the historical features (\hf) and the environment features (\ef); as well as the \rew{}, including the main attributes used for its computation.\footnote{The complete MDP description is available at \url{tinyurl.com/2p8aytrb}.}

For example, Table~\ref{tab:preprocessing} shows how a trace related to the \fines{} scenario is preprocessed and transformed into an annotated trace, and then into MDP actions, states and rewards. 
The trace activities are annotated according to whether they have been carried out  either by the \emph{agent} or by the \emph{environment}, and the attributes \attr{2months} (the bimester since the fine creation), \attr{amClass} (the fine amount class) and \attr{payType} (type of payment performed) are computed. In the MDP construction step, the agent's activities (with the bimester interval
\footnote{\label{note1}The MDP actions in this scenario take into account, besides the activity name, also the 2-month interval (since the creation of the fine) in which the activity has been carried out (\attr{2months}).}) are used as actions, while the state is built by leveraging the last executed activity (\la), the \attr{2months} and the \attr{amClass} attributes. The reward is not null when the payment is finally received and since in this trace the full payment is received after $6$ months, $2$ credits are awarded.

Once the log is enriched it is passed to the MDP generation step. We build two MDPs: the MDP$_{\bpi}$ for the \loanreq scenario (with 982 states and 15 actions) and the MDP$_{\finel}$ for the \traffine scenario (with 215 sates and 70 actions).

\begin{table}[t]
\centering
\scalebox{0.8}{
    \begin{tabular}{l@{\hskip 0.2in} l@{\hskip 0.2in} l@{\hskip 0.1in} l@{\hskip 0.1in} l@{\hskip 0.2in} }
		\toprule
			\textbf{Scenario} & \multicolumn{4}{l}{\textbf{MDP description}}  \\ \midrule
			\multirow{16}{*}{\loans} & \multirow{2}{*}{\action} & \multicolumn{3}{l}{Bank activities: loan acceptance, loan  rejection, offer creation and delivery, requests for} \\  
			& & \multicolumn{3}{l}{customer response}\\ \cline{2-5} 
			
			& \multirow{8}{*}{\state} & \multirow{2}{*}{\la} & \multicolumn{2}{l}{last activity of the agent (bank) or of the environment (customer)}\\
			& & & \multicolumn{2}{l}{Customer activities: application cancellation, offer sent back to the bank, offer acceptance}\\ \cline{3-5}
			& & \multirow{5}{*}{\hf} &  \attr{call\#} & \# of bank calls after the offer is sent \\
			& & & \attr{miss\#} & \# of requests for missing information\\ 
			& & & \attr{offer\#} & \# of offers to the customer\\
            & & & \attr{reply\#} & \# of customer replies to the offer \\			
			& & &\attr{fix} & true if wrong inputs in the application are fixed\\  \cline{3-5}
			&  & \multirow{1}{*}{\ef} & \attr{amClass} & loan amount class: low ($\le 6000$), medium and high ($> 15000$) \\ \cline{2-5}
			
			& \multirow{3}{*}{\rew} & \multirow{3}{*}{\emph{attr.}}  &\attr{duration} & activity average duration \\ 
			& & & \attr{amClass} & loan amount class \\
			& & & \attr{granted} & whether the loan has been granted \\ \cline{3-5}
			& &\multicolumn{3}{l}{The reward is computed for each MDP state so that the reward of the complete}\\
			& & \multicolumn{3}{l}{execution corresponds to the value of the $\text{KPI}_\bpi$ for that execution.\tablefootnote{The component of the reward for an MDP state $s$ related to the interest of the bank is multiplied by a coefficient $c(n)=\frac{(n/\lambda)^2}{1+(n/\lambda)^2}$ that depends on the number of occurrences $n$ of the event log traces that pass through the specific MDP edge with outgoing state $s$. $c$ goes to 1 when $n$ grows. Here $\lambda$ is a parameter that can be opportunely tuned, we selected $\lambda=3$ which corresponds to the median number of edge occurrences in the MDP. This factor is needed to discourage during the RL training the exploitation of some actions that have a positive reward but have low statistic reliability.}} \\ \hline

			\multirow{10}{*}{\fines} &
			 \multirow{2}{*}{\action
			 \footnotemark[\getrefnumber{note1}]
			 } & \multicolumn{3}{l}{Police department activities: fine creation and delivery, penalty increase}\\ 
			 & & \multicolumn{3}{l}{and request for credit collection }\\ \cline{2-5}

			& \multirow{4}{*}{\state} & \multirow{2}{*}{\la} &  \multicolumn{2}{l}{last activity of the agent (police department) or of the environment (offender)} \\ 
			& & & \multicolumn{2}{l}{Offender activities: appeal to the Prefecture or to the Judge, payment}\\ \cline{3-5}
			& & \hf & \attr{2months} & number of two-month intervals since the creation of the fine \\ \cline{3-5}
			& & \multirow{1}{*}{\ef}& \attr{amClass} & fine amount class: low (amount $< 50$), high (amount $\ge 50$). \\ 
            \cline{2-5}
           
            & \multirow{4}{*}{\rew} & \multirow{2}{*}{\emph{attr.}} & \attr{2months} & number of two-month intervals since the creation of the fine\\
            & & & \attr{payType} & type of payment (null, partial, full or appeal)\\ \cline{3-5}
            & & \multicolumn{3}{l}{The reward is computed for each MDP’s state so that the reward of the complete}\\
            & & \multicolumn{3}{l}{execution corresponds to the value of the $\text{KPI}_{\finel}$ for that execution.}\\ 
		\bottomrule
		\end{tabular}
}
\vspace{0.3cm}
\caption{MDP for the \loanreq and the \traffine scenarios.}
\vspace{-0.5cm}
\label{tab:scenarios}
\end{table}

\begin{table}[t]
	\centering
	\scalebox{0.7}{%
\begin{tabular}{l@{\hskip 0.1in} c@{\hskip 0.1in} c@{\hskip 0.3in} c@{\hskip 0.1in} c@{\hskip 0.1in} c@{\hskip 0.1in} c@{\hskip 0.3in} l@{\hskip 0.1in} l@{\hskip 0.1in} c}
\toprule
\multicolumn{3}{l}{\textbf{Trace}} & \multicolumn{4}{l}{\textbf{Enriched trace}} & \multicolumn{3}{l}{\textbf{MDP}} \\
\textbf{activity} & \textbf{timestamp} & \attr{amount}
 & \attr{2months} & \attr{amClass} & actor & \attr{payType} & \action & next \state & \rew \\ \hline
\act{Create fine} & 13/1/21 & 40 & 0 & low & agent & - & \act{Create fine-0} &  $\langle \text{\act{Create fine}, 0, low} \rangle$ & 0 \\
\act{Send fine} & 24/1/21 & 40 & 0 & low & agent & - & \act{Send fine-0} & $\langle \text{\act{Send fine}, 0, low} \rangle$ & 0\\
\act{Add penalty} & 18/3/21 & 60 & 1 & high & agent & - & \act{Add penalty-1} & - & -\\
\act{Payment} & 25/7/21 & 60 & 3 & high & env. & full & - & $\langle \text{\act{Payment}, 3, high} \rangle$ & 2 \\
\bottomrule
\end{tabular}
}
	\vspace{0.3cm}
	\caption{Example of the transformation of a trace in the corresponding MDP components.}
	\vspace{-1cm}
	\label{tab:preprocessing}
\end{table}

\subsection{Experimental Setting}
\label{ssec:setting}

In order to answer our research questions, the two event logs have been split in a training part, which is used in in the \emph{RL phase}, and a test part, which is used for the evaluation of the learned policy. For evaluating the computed policies, since in this setting both training and test set size can impact the evaluation results, we use two different splitting criteria (defining the percentage of event log used for the training and the test set): (i) \splitA{} (60\% of the traces for the training set and 40\% for the test set) and (ii) \splitB{} (80\% for the training and 20\% for the testing).
%
%
For the evaluation of the optimal policy obtained by RL and for answering our two research questions, two different evaluations have been carried out: a \simeval and a \logeval.

The \simeval uses a Monte Carlo simulation similar to the one used in the training phase, but, differently from the training phase, where the MDP is obtained from the training log, here a test MDP, obtained from the test log, is leveraged to simulate the environment response. In this simulation, the optimal policy obtained from the RL approach is compared, in terms of the KPI of interest, against a random policy and against a policy
corresponding to the most frequent decisions made by the actor in the actual traces.
The value of the reward for each of the simulated policy is computed as the average over 100.000 simulated cases. This evaluation provides a preliminary answer to the first research question \ref{RQ1}.



The \logeval aims at comparing the optimal policy obtained from RL with the actual policies used in the process. It is used  for answering both our research questions. 
For \ref{RQ1}, we focus on the policy recommended when no activity has been executed yet. In this setting, we compare the value of the KPI of interest for the traces in the test event log that follow the optimal policy (from the first activity) (i) with the value of the KPI of interest of all the traces in the event log, and (ii) with the value of the KPI of interest of the traces in the event log that do not follow the recommended optimal policy.
For \ref{RQ2}, we focus on the policy recommended for ongoing executions, i.e., when some activity has already been executed. We hence consider, for each trace in the test event log, all its prefixes and separately analyze each of them, as a potential ongoing execution. For each prefix $p$ of a trace $t$ in the test event log we compare the value of the KPI of interest of the trace $t$ once completed against an estimation of the value of the KPI obtained following the optimal policy from that execution point forward. The estimation is obtained by averaging the KPI values of the traces in the log that have the same prefix as the reference prefix $p$ and follow the optimal policy from there on. 



\subsection{Results}
\label{ssec:results}
In this section we report the results of the two scenarios related to the event logs described in Section~\ref{ssec:datasets}. For both scenarios, as described in Section~\ref{ssec:setting}, we show (i) the results related to the evaluation of the complete optimal policy (\ref{RQ1}) by reporting first the \simeval and then the \logeval; and (ii) the results related to the evaluation of the optimal policy on the test log assuming that some events have already been executed (\ref{RQ2}).

\vspace{0.15cm}
\noindent \emph{Research Question \ref{RQ1}.}
Table~\ref{tab:simulator} reports the results related to the \simeval for both the \loanreq and the \traffine scenarios. 
For both splitting criterion (\splitA{} and \splitB) and for each policy analysed,
the average KPI value is displayed together with
the percentage of executions for which the bank offer has been accepted by the customer (or the fines have been fully paid by the offender).
The policies analysed are: the random policy (\random), the policy selecting the most frequent action in the log for each state
(\usual) and the optimal (\optimal) policy.

\begin{table}[t]
	\centering
 	\scalebox{0.8}{%
		\begin{tabular}{l@{\hskip 0.2in} l@{\hskip 0.2in} l@{\hskip 0.2in} c@{\hskip 0.2in} c}
		\toprule
			\multirow{2}{*}{\textbf{Scenario}} & \textbf{Splitting} & \multirow{2}{*}{\textbf{Policy}} & \multirow{2}{*}{\textbf{avg. KPI}} & \textbf{\act{Offer accepted} / 
			}\\
			& \textbf{criterion} & & & \textbf{full \act{Payment} 
			}\\\hline
			\multirow{6}{*}{\loans} & \multirow{3}{*}{\splitA} & \random & $36.8$ & $1.4\%$ \\
			& & \usual & $1497.5$ & $38.9\%$ \\
			& & \optimal & $1727$ & $53.7\%$ \\ \cline{2-5}
			& \multirow{3}{*}{\splitB} & \random & $35.7$ & $1.5\%$ \\
			& & \usual & $1710.5$ & $43.5\%$ \\
			& & \optimal & $1965.1$ & $61.7\%$ \\ \hline
			\multirow{6}{*}{\fines} & \multirow{3}{*}{\splitA} & \random & $1.02$ & $36.9\%$ \\
			& & \usual & $1.12$ & $41.7\%$ \\
			& &\optimal & $1.17$ & $42.3\%$ \\ \cline{2-5}
			& \multirow{3}{*}{\splitB} & \random & $0.84$ & $30.0\%$ \\
			& & \usual & $0.97$ & $35.0\%$ \\
			& & \optimal & $1.05$ & $37.1\%$ \\	
		\bottomrule
		\end{tabular}
 		}
	\vspace{0.3cm}
	\caption{Results of the \simeval for the \loanreq and the \traffine scenarios.}
	\vspace{-1cm}
	\label{tab:simulator}
\end{table}


The rows related to the \loanreq scenario (\loans) show that for both splitting criteria, the optimal policy (\optimal) generates an average KPI value
much higher than the one obtained with a random policy (\random), but also  higher than the one obtained with a policy characterized by frequently taken actions (\usual). This result confirms that the proposed \optimal policy actually outperforms the policy that is frequently taken in the actual traces, which is considered to be an ``optimal'' policy by the target agent.
Different optimal (and \usual) policies are returned with different splitting criteria. When learning with a larger training set and simulating on a smaller test set, the average KPI value increases, for the \optimal and the \usual policy, while slightly decreases for the \random policy.
Moreover, the table also shows the percentage of traces that, based on the policy simulations, are finally accepted by the customer. By changing the data splitting criteria, the effect is similar to that observed for the average KPI value for the \optimal and the \usual policy, with a percentage of accepted offers raising from around $39\%$ to $43\%$ for the \usual policy and from around $53\%$ to more than $60\%$ with the \optimal policy. An almost null increase is observed instead for the \random policy.


The results related to the \traffine scenario are similar to the results of the loan scenario, as shown in the row \fines{} of Table~\ref{tab:simulator}. As for the loan scenario, also in this case, for both splitting criteria, the optimal policy returns higher average KPI values
(and hence lower money waste) and produces a higher percentage of traces with fully paid fines than the \random and the \usual policies. Also in this case, the difference between the \optimal and the \usual policy confirms that the proposed recommendation policy improves the policy actually used in practice. In this scenario, however, the difference in terms of percentage of traces for which fines have fully been paid between the optimal and the random policy is lower than for the \loanreq case. This is possibly due 
to the overall higher percentage of traces in the \finel{} event log for which the fines have been fully paid (40\%)  with respect to the percentage of traces in the \bpi{} log for which the loan offer has been accepted by the customer (17\%), as well as to the higher number of actions of MDP$_{\finel}$ with respect to the number of actions of MDP$_{\bpi}$. Moreover, differently from the \loanreq{} scenario, there is an overall decrease in terms of average KPI value and of traces with fully paid fines when using a larger training set and a smaller test set (\splitB{} splitting criterion).


Table~\ref{tab:log_analysis} shows the results related to the \logeval. For each of the two scenarios and for each splitting criterion, we report the number of traces, the average KPI value, 
as well as the percentage of traces for which the offer has been accepted (or the fines have been fully paid) for (i) all traces in the test set (\allt), (ii) the traces in the test set that follow the optimal policy (\optimalt); (iii) the traces in the test set that do not follow the optimal policy (\notoptimalt).

\begin{table}[t]
	\centering
 	\scalebox{0.75}{%
		\begin{tabular}{l@{\hskip 0.2in} l@{\hskip 0.2in} l@{\hskip 0.2in} c@{\hskip 0.2in} c@{\hskip 0.2in} c}
		\toprule
			\multirow{2}{*}{\textbf{Scenario}} & \textbf{Splitting} & \multirow{2}{*}{\textbf{traces}} & \multirow{2}{*}{\textbf{trace \#}}  & \textbf{avg} &  \textbf{\act{Offer accepted} / 
			} \\ 
			& \textbf{criterion} & & & \textbf{KPI} & \textbf{full \act{Payment} 
			}\\ \hline
			\multirow{6}{*}{\loans} & \multirow{3}{*}{\splitA} & \allt & $5197$ & $583.3$ & $16.1\%$ \\
			& & \optimalt & $1384$ ($26.6\%$) & $1249.7$ & $34.4\%$ \\
			& & \notoptimalt & $3813$ ($73.4\%$) & $341.5$ & $9.4\%$ \\ \cline{2-6}
			& \multirow{3}{*}{\splitB} & \allt & $2600$ & $537.2$ & $14.3\%$ \\
			& & \optimalt & $753$ ($29\%$) & $1082.2$  & $30.4\%$ \\
			& & \notoptimalt & $1847$ ($71\%$) & $315.1$ & $7.7\%$ \\ \hline
			\multirow{6}{*}{\fines} & \multirow{3}{*}{\splitA} & \allt & $59946$ & $1.11$ & $40.5\%$ \\
			& & \optimalt & $22665$ ($37.8\%$) & $2.68$ & $90.9\%$ \\
			& & \notoptimalt & $37281$ ($62.2\%$) & $0.15$ & $9.9\%$ \\ \cline{2-6}
			& \multirow{3}{*}{\splitB} & \allt & $29973$ & $0.96$ & $34.7\%$ \\
			& & \optimalt & $9243$ ($30.8\%$) & $2.76$ & $92.7\%$ \\
			& & \notoptimalt & $20730$ ($69.2\%$) & $0.16$ & $8.9\%$ \\
		\bottomrule
		\end{tabular}
		}
	\vspace{0.3cm}
	\caption{Results related to the \logeval for the \loanreq and the \traffine scenario.}
	\vspace{-1cm}
	\label{tab:log_analysis}
\end{table}

The results of the \logeval for the \loanreq scenario (
\loans) confirm the results obtained with the \simeval. For both splitting criteria, indeed, the average KPI value 
of the traces following the optimal policy (\optimalt) is higher than the average 
KPI value of all the traces (\allt), which in turn is higher than the average KPI value 
of the traces that do not follow the optimal policy (\notoptimalt).
The traces following the optimal policy generate an average bank profit of more than 500 euros higher than the average bank profit of all the traces in the event log, as well as of more than 750 euros higher than the average bank profit of the traces that do not follow the optimal policy.
The same ranking is obtained if the compared approaches are ordered by the percentage of traces for which the 
offer by the bank has been accepted by the customer: around 30\% for the traces following the optimal policy, around 15\% for all traces, and less than 10\% for the traces not following the optimal policy. 
No major differences can be observed
between the two splitting criteria, except for a small decrease of the average KPI value and of
the percentage of accepted offers.


Similarly to the \loanreq scenario, also in the \traffine scenario (rows \fines{} in Table~\ref{tab:log_analysis}) the results of the \logeval confirm the findings of the \simeval. Indeed, for both splitting criteria, the traces following the optimal policy (\optimalt) obtain an average KPI value 
higher than the average KPI value 
of all the traces (\allt), which in turn is higher than the average KPI value 
of the traces that do not follow the optimal policy (\notoptimalt). The traces following the optimal policy can produce an average credit value of more than 1 credit higher than the average credit value of all the traces in the event log, as well as of more than 2 credits higher than the average credit value of the traces that do not follow the optimal policy. The trend is also similar for the percentage of traces for which the fine is fully paid. Around 90\% of the traces that follow the optimal policy are able to get fully paid fines for both the splitting criteria. While, as in the \loanreq, the percentage of traces with a fully paid fine decreases from the 40\% test event log to the 20\% event log for the \allt and \notoptimalt policies, for \optimalt the percentage of traces for which the full payment is received is higher for the 20\% than for the  40\% test event log.

The above results of the two scenarios clearly show that, when no activity has been executed before the target agent starts following the recommendations, the sequence of next activities suggested by the optimal policy generates an average 
value for the KPI of interest
higher than a random policy
and than a policy
following the most frequently taken actions
and, on average, higher than the average KPI value obtained by the actual executions in the test event log (\ref{RQ1}). No clear trends can be observed for different splitting criteria.



\begin{figure}[t]
\centering
\subfloat[average delta KPI value]{\label{sublabel:avg_loan}
  \includegraphics[width=0.5\textwidth]{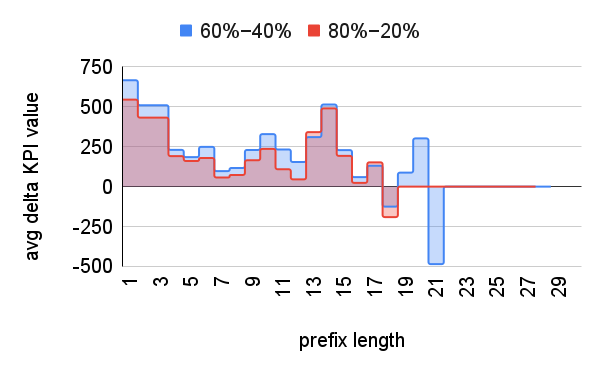}}
\subfloat[number of traces for each prefix]{\label{sublabel:tot_loan}
  \includegraphics[width=0.5\textwidth]{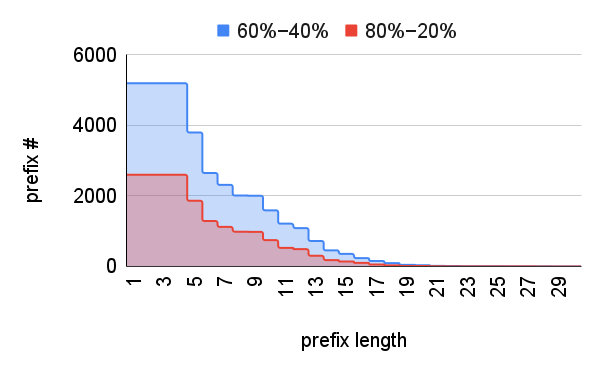}}
\caption{Prefix analysis for the \loanreq scenario.} 
\vspace{-0.5cm}
\label{fig:loan_prefix}
\end{figure}

\begin{figure}[t]
\centering
\subfloat[average delta KPI value]{\label{sublabel:avg_fines}
  \includegraphics[width=0.5\textwidth]{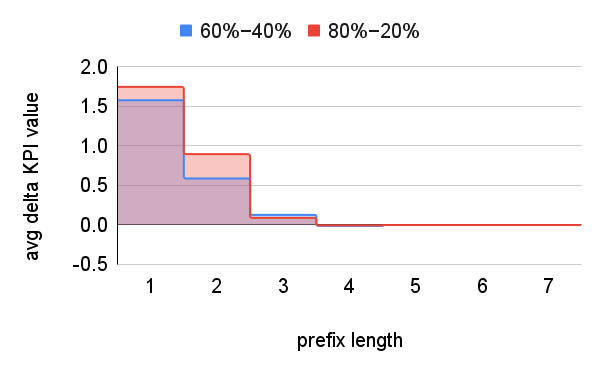}}
\subfloat[number of traces for each prefix]{\label{sublabel:tot_fines}
  \includegraphics[width=0.5\textwidth]{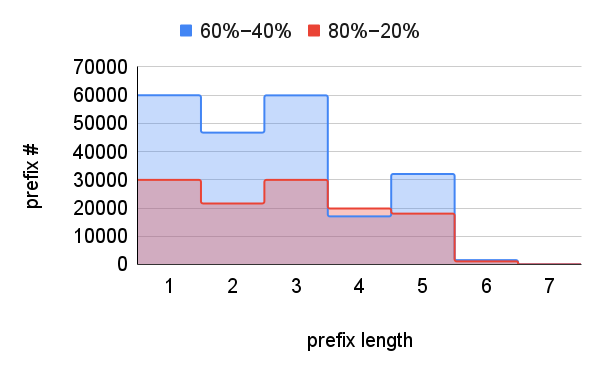}}
\caption{Prefix analysis for the \traffine scenario.} 
\vspace{-0.5cm}
\label{fig:fine_prefix}
\end{figure}

\vspace{0.15cm}
\noindent \emph{Research Question \ref{RQ2}.}
As described in Section~\ref{ssec:setting}, we also evaluate the optimal policies at different prefix lengths, that is, by assuming that a part of the execution  has already been carried out, before the target agent starts adopting the optimal policy. \figurename~\ref{fig:loan_prefix} and \figurename~\ref{fig:fine_prefix} show the average delta KPI value for each prefix length, as well as the prefix occurrence per prefix length. The delta KPI value for each trace and prefix length is computed as the difference between the KPI value obtained by following the optimal policy from that prefix on and the KPI value of the complete trace related to that prefix.

The plot corresponding to the \loanreq scenario (\figurename~\ref{sublabel:avg_loan}) shows that for both splitting criteria and for prefix lengths up to $18$ there is an average positive delta KPI value,  
while for longer prefixes a negative or almost null average KPI values are observed. 
These results can be explained by the low number of traces with length higher than $18$ in the test event logs, as it is shown 
in \figurename~\ref{sublabel:tot_loan}.  

In the \traffine scenario, the plot in \figurename~\ref{sublabel:avg_fines} shows a relatively high delta average KPI value 
for short prefixes (prefixes of length 1 and 2), while the average delta KPI value 
starts decreasing for traces of prefix length 3.  
Also in this case, as for the other 
scenario, the decrease in terms of delta KPI value 
is mainly due to an overall decrease of the number of traces after prefix $3$ (see \figurename~\ref{sublabel:tot_fines}). 
Differently from the \loanreq scenario, as already observed
during the discussion of \ref{RQ1},
the average delta KPI value 
obtained with the \splitB{} splitting criterion is higher than the one 
obtained with 
the \splitA{} splitting criterion, except that for prefix length $3$. 

In conclusion, these results confirm that even when considering ongoing executions, the recommended sequence of next activities suggested by the proposed optimal policy generates higher average KPI values 
than the ones obtained by actual executions in the test event log (\ref{RQ2}).

Beyond the performance perspective, we briefly comment here on the plausibility of the optimal policies obtained. The major contributions of the policies for the two cases are clear and reasonable. In the \loanreq scenario the policy advises to accept more loan applications, so as to increase the number of possible accepted loans. Moreover, it advises to increase the interaction between the bank and the customer, with the creation of multiple offers and the subsequent call to the customer.
In the \traffine scenario the policy advises to send the fine early to the offender, so as to raise the probability that he/she pays the fine on time.



\section{Related Work}
\label{sec:related}
The state-of-the-art works related to this paper pertain to two fields: Prescriptive Process Monitoring and Reinforcement Learning. The section is hence structured by first presenting Prescriptive Process Monitoring related works and then Reinforcement Learning state-of-the-art works, applied to process mining problems.   

Several Prescriptive Process Monitoring techniques have been recently proposed in the literature. 
Focusing on the type of interventions that the approaches recommend~\cite{DBLP:journals/corr/abs-2112-01769}, we can roughly classify existing work in Prescriptive Process Monitoring in three main groups: (i) those that recommend different types of interventions to prevent or mitigate the occurrence of an undesired outcome~\cite{Teinemaaetal2018,fahrenkrog2019fire,metzger2019proactive,metzger2020triggering,DBLP:journals/corr/abs-2109-02894}; (ii) those that take a resource perspective and recommend a resource allocation~\cite{10.1007/978-3-319-39696-5_35,DBLP:conf/icpm/ParkS19}; (iii) those that provide recommendations related to the next activity to optimize a given KPI~\cite{weinzierl2020predictive,groger2014prescriptive,de2020design}.



The approach presented in this paper falls under this third family of  prescriptive process monitoring approaches. Only a small amount of research has been done in this third group of works.
Weinzierl et al. in~\cite{weinzierl2020predictive} discuss how the most likely behavior does not guarantee to achieve the desired business goal. As a solution to this problem, they propose and evaluate a prescriptive business process monitoring technique that recommends next best actions to optimize a specific KPI, i.e., the time. Gröger et al. in~\cite{groger2014prescriptive} present a data-mining driven concept of recommendation-based business process optimization supporting adaptive and continuously optimized business processes. 
De Leoni et al. in~\cite{de2020design} discuss Process-aware Recommender (PAR) systems, in which a prescriptive-analytics component, in case of executions with a negative outcome prediction, recommends the next activities that minimize the risk to complete the process execution with a negative outcome. 
%
Differently from these state-of-the-art works, however, in this work we take the perspective of one of the actors of the process and we aim at optimizing a domain-specific KPI of interest for this actor by leveraging an RL approach.

In the literature, only few RL approaches have been proposed for facing problems in the process mining field. Silvander proposes using Q-Learning with function approximation via a deep neural network (DQN) for the optimization of business processes~\cite{silvander2019business}. He suggests defining a so called decay rate to reduce the amount of exploration over time. 
Huang et al. employ RL for the dynamic optimization of resource allocation in business process executions~\cite{huang2011reinforcement}. 
Metzger et al. propose an alarm-based approach to prevent and mitigate an undesired outcome~\cite{metzger2020triggering}. They use online RL to learn when to trigger proactive process adaptations based on the reliability of predictions. 
Although all these works use RL in the process mining field, none of them use it for recommending the next actions to perform in order to optimize a certain KPI of interest, as in this work.

Finally, some works have applied RL and Inverse Reinforcement Learning (IRL) approaches to recommend the next actions on temporal data~\cite{DBLP:conf/recsys/Massimo018} or on data constrained by temporal constraints~\cite{DBLP:conf/aips/GiacomoIFP19}.

\section{Conclusion}
\label{sec:conclusion}
In this paper we have proposed the use of RL in the solution of the problem of computing next activity recommendations in Prescriptive Process Monitoring problems.

Differently from other state-of-the-art works our model handles non deterministic  processes, in which only part of the activities are actually actionable and the rest of them are, from the target actor point of view, stochastically selected by the system environment. This is a common situation in multi-actors processes.
By taking the decision making perspective of one of the actors  involved in a process (target actor), we first learn from past executions the behaviour of the environment and we then use RL to recommend the best activities to carry on in order to optimize a measure of interest. The obtained results show the goodness of the proposed approach in comparison to the policy used by the actor, i.e., without using recommendations.

We plan to extend this approach by including in the MDP state the raw information related to the history of the process execution, so as to automate as much as possible the pre-processing phase of our computational pipeline. However, in that case the consequent increase of the state space dimension and its cardinality  would require the usage of state generalisation techniques, such as, those implemented with Deep Reinforcement Learning or by applying smart clustering techniques. Moreover, we would like to explore the possibility to use declarative constraints for defining and enforcing domain knowledge constraints.


%
%
%
\bibliographystyle{splncs04}
\bibliography{bibliography}
\end{document}